\documentclass[a4paper,fleqn]{cas-sc}

\usepackage[sort&compress,numbers]{natbib}

\usepackage[numbers]{natbib}
\usepackage{framed,multirow}
\usepackage{amsmath}

\usepackage{soul}
\usepackage{chngcntr}

\usepackage{amssymb}
\usepackage{latexsym}
\usepackage{amsmath}
\usepackage{nccmath}
\usepackage{color,soul}
\usepackage[linesnumbered,ruled]{algorithm2e}

\usepackage{url}
\usepackage{xcolor}
\usepackage{multirow}
\usepackage{array}
\usepackage{graphicx}

\def\tsc#1{\csdef{#1}{\textsc{\lowercase{#1}}\xspace}}
\tsc{WGM}
\tsc{QE}
\tsc{EP}
\tsc{PMS}
\tsc{BEC}
\tsc{DE}

\begin{document}
\let\WriteBookmarks\relax
\def\floatpagepagefraction{1}
\def\textpagefraction{.001}
\shorttitle{A hybrid classification-regression approach for 3D hand pose estimation using graph convolutional networks}
\shortauthors{I Kourbane et~al.}
\title [mode = title]{A hybrid classification-regression approach for 3D hand pose estimation using graph convolutional networks}                      
 \author[1]{Ikram Kourbane}[orcid=0000-0001-8753-6710]
  \cormark[1] 
  \ead{ikourbane@gtu.edu.tr} 
   \author[1]{Yakup Genc}
  \address[1]{Gebze Technical University, Gebze, Kocaeli, 41000, Turkey}
  \cortext[cor1]{Corresponding author.} 

\begin{abstract}
Hand pose estimation is a crucial part of a wide range of augmented reality and human-computer interaction applications. Predicting the 3D hand pose from a single RGB image is challenging due to occlusion and depth ambiguities. GCN-based (Graph Convolutional Networks) methods exploit the structural relationship similarity between graphs and hand joints to model kinematic dependencies between joints. These techniques use predefined or global learned joint relationships, which may fail to capture pose dependent constraints. To address this problem, we propose a two-stage GCN-based framework that learns per-pose relationship constraints. Specifically, the first phase quantizes the 2D/3D space to classify the joints into 2D/3D blocks based on their locality. This spatial dependency information guides this phase to estimate reliable 2D and 3D poses. The second stage further improves the 3D estimation through a GCN-based module that uses an adaptative nearest neighbor algorithm to determine joint relationships. Extensive experiments show that our multi-stage GCN approach yields an efficient model that produces accurate 2D/3D hand poses and outperforms the state-of-the-art on two public datasets.
\end{abstract}




\maketitle

\section{Introduction}
\label{introduction}

Hand pose estimation is one of the well-known and fast-growing topics in the computer vision community. It plays a significant role in multiple application domains, such as human-computer interaction, augmented reality, virtual reality and robotics. Although there is a large body of study in the literature, solving the 3D hand pose estimation problem remains challenging due to the similarity among fingers, self-occlusion and the complexity of the hand poses.

With the introduction of affordable commodity depth sensors and the rapid development of deep learning techniques, RGBD-based approaches achieve accurate 3D hand pose estimation results \cite{signal1, depth1,depth3,pixel,signal3,point,voxel}. Other methods employ calibrated multi-view cameras to reduce the depth ambiguity \cite{multi1, multi2}. However, both depth and multi-view solutions are not readily available and work only in constrained environments. RGB-based methods are preferable since they are more accessible and do not require controlled settings. Due to the absence of depth information and occlusion, this task becomes an ill-posed problem. Furthermore, collecting 3D annotations for real images requires complex multi-view setups and manual annotations. 

Learning-based methods tackle RGB-based 3D hand pose estimation by creating effective models trained on large-scale datasets \cite{mueller,gomez1,fre}. We can divide them mainly into two categories. The first one adopts generative models to find the closet configuration (model-based) \cite{mueller,spurr,yang,gu}. The second category uses discriminative models to extract features and estimate the 3D pose (appearance-based) \cite{panteleris, zimmermann, cai,iqbal,boukhayma, zhou}. Despite the promising results, most of them assume that the model can implicitly learn kinematic relationships between the joints. However, this task is challenging and requires explicit constraints to guide the model optimization. 

Inspired by the natural graph representation of the hand, recent studies use graph convolution networks (GCN) \cite{graph} to model skeletal constraints between joints \cite{ge,guo,cai2,semantic,hope}. \cite{cai2} exploits annotated video frames to enforce spatial and temporal relationships between the joints based on predefined semantic meanings. This approach is limited to consecutive images, where temporal information is present. \cite{semantic} estimates the 3D human pose by learning weighted relationships between body joints based on a predefined human skeletal. However, this approach misses potential relationships of physically disconnected joints. \cite{hope} addresses this limitation by replacing the skeletal relationships with global learnable constraints between all hand poses. But, different hand poses should not share the same relationship constraints since they do not have the same joint configurations (e.g: open and closed hands).

\begin{figure*}
\centering
\includegraphics[width=0.99\textwidth, height=0.335\textheight]{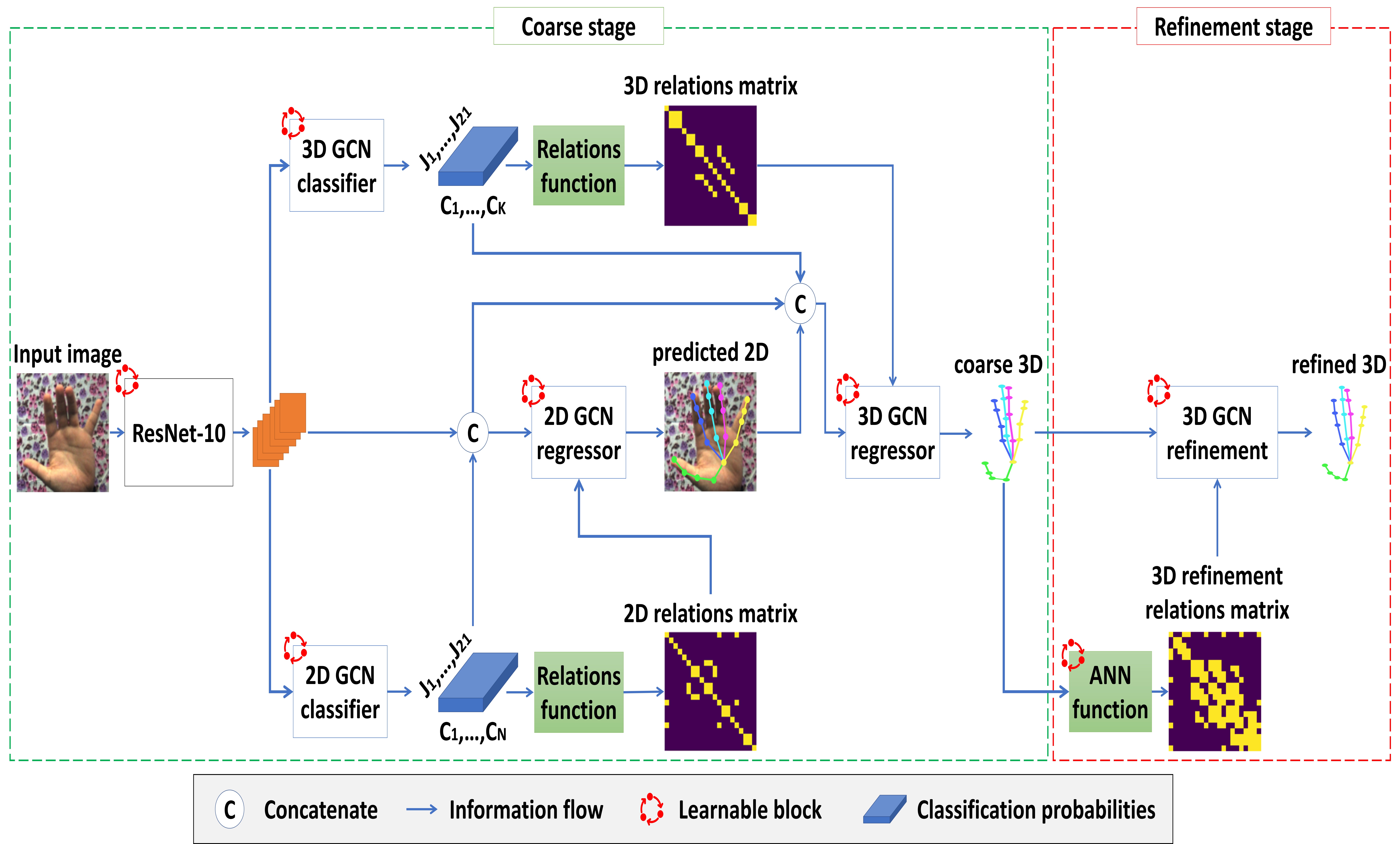}
\caption{Schematic overview of our 3D pose estimation framework, where the input is the RGB hand image, and the output is the 3D pose. We extract image feature maps using a lightweight ResNet-10 network. We classify joints (i.e., $J_i$) in 2D and 3D spaces based on their spatial alignment such that neighboring joints belong to the same class $C_i$, where $N$ and $K$ are the number of classes in 2D and 3D spaces, respectively. We construct 2D/3D per-pose joint relationship constraints using 2D/3D GCN-based classifiers and \textit {Relations} function. This knowledge guides 2D/3D GCN-based regressors to estimate reliable 2D/3D poses. We further improve the 3D pose using the refinement network that employs an adaptative nearest neighbors (ANN) function to identify joint relationships (This figure should be printed in color).}
\label{fig:architecture}
\vspace{-3mm}
\end{figure*}

Motivated by these observations, we introduce a two-stage GCN-based framework to estimate 3D hand pose from a single RGB image. Unlike previous methods that either employ predefined joint relationship constraints \cite{cai2, semantic, ge} or learn a global one for the entire dataset \cite{hope}, we learn per-pose relationship constraints expressing global and local joint spatial dependencies. Specifically, we propose a hybrid classification-regression framework that employs a light-weight feature extractor (ResNet-10) \cite{resnet} and GCN modules \cite{graph} to classify each 2D/3D joint into its corresponding 2D/3D blocks based on its spatial locality. In particular, we divide our 2D/3D space into multiple classes such that neighboring joints belong to the same class. We utilize the classification probabilities to construct per-pose geometric constraints for 2D/3D joints. Afterward, we regress an initial (coarse) 3D hand pose using another GCN module that exploits learned constraints to guide the pose estimation task. (Fig.~\ref{fig:architecture}). Finally, we further improve the coarse pose using ANN-based (Adaptative Nearest Neighbors) GCN. Specifically, instead of fixing K neighbors for all the nodes (KNN), we calculate the distance between all joints to learn a differentiable threshold that determines the number of neighbors of each node in the graph. The proposed approach speeds up the running time since it includes lightweight feature extractor \cite{resnet} and GCN-based modules \cite{graph}. The latter proved to be computationally efficient than CNNs \cite{hope}. We summarize our contributions as follows: 

\begin{itemize}
\item We propose a two-stage GCN-based approach that combines classification and regression to estimate an accurate 3D hand pose from a single RGB image. The classifiers find spatial dependencies between neighboring joints to learn per-pose relationship constraints that guide the regression.
\item Instead of KNN-based GCN, we propose an adaptative nearest neighbor algorithm that learns a different number of joint relationships improving the 3D hand pose estimation task.
\item To evaluate our approach, we conduct extensive experiments on synthetic and real-world datasets. The reported quantitative and qualitative results demonstrate that our efficient method outperforms state-of-the-art and estimate accurate 2D/3D hand poses.
\end{itemize}

The rest of this paper is organized as follows: Section.~\ref{related_works} reviews the related studies of our work. Section.~\ref{methodology} explains the proposed method in-depth and describes the most important modules of our framework. Section.~\ref{experimental} presents experimental results on two public datasets and compares the proposed approach against the state-of-the-art. Finally, Section.~\ref{conclusion} presents the conclusion of the study and direction for future work.

\section{Related works}
\label{related_works}
We classify 3D hand pose estimation methods based on the input modality into three categories depth-based, multiview RGB-based, and monocular RGB-based. In the past few years, numerous studies adopt deep learning techniques to handle various types of inputs for depth methods, such as 2D pixels \cite{signal1,depth1,pixel,signal3}, a set of 3D points \cite{point, depth3} or voxels \cite{voxel}. In the second category, several studies use many RGB cameras on different angles to alleviate the occlusion problem \cite{multi1,multi2}.

RGB cameras are more available compared to the two-mentioned categories. There are several attempts to solve the problem with model-based methods \cite{chpr,icppso,pso}. These methods search the closest configuration that fits the hand model via an optimization process. They require powerful prior knowledge about many dynamic and physics hypotheses, which leads to poor performances. Deep generative-based approaches exhibit promising performances and generate accurate 3D poses. \cite{mueller} employs generative adversarial networks (GANs) \cite{gan} to translate synthetic labeled data to realistic images to reduce the domain gap. \cite{spurr} uses Variational Autoencoders (VAEs) \cite{vae} to learn a shared latent space across RGB and depth modalities to remedy the missing depth ambiguity. However, These methods model black-box latent representations, which synthesize a single 3D pose for a given RGB image. To address this issue, \cite{yang,gu} disentangle the diverse factors that affect the hand visualization, including camera viewpoint, scene context and background.

Discriminative methods act as a function that maps directly between the RGB image and the 3D pose using large-scale annotated datasets. \cite {zimmermann} is one of the first attempts that estimates the 3D hand pose from a monocular RGB image. It employs synthetic data and three concatenated networks: hand segmenter, 2D keypoint and 3D hand pose predictors. After this work, many deep-based studies have been proposed to improve 3D hand pose estimation. \cite{panteleris} estimates 3D hand poses by combining CNN with a kinematic 3D hand model. \cite{cai} leverage depth information to supervise the training by having a pair of RGB-RGBD images. \cite{iqbal} uses a two-stack hourglass with latent 2.5D heatmaps and lift them to 3D. Other approaches achieve reliable 3D poses by regressing the hand mesh and estimating the 3D pose from it \cite{boukhayma, baek, zhou, zhang}. However, these methods require additional supervision since the datasets must include both the meshes and the 3D poses. More recently, \cite{zhao} proposed a knowledge distillation and generalization framework for RGB-based pose estimation. 

GCN \cite{graph} became a hot topic of research in the computer vision field since there exist several problems that have graph-like structures as input. \cite{ge} applies GCNs \cite{graph} to estimate the 3D pose with the help of a weakly-supervised training strategy that employs depth regularizer. \cite{cai2} exploits spatial and temporal relationships for 3D human and hand pose estimation tasks. \cite{guo} introduces a self-supervised module that uses 2D relationships and 3D geometric knowledge to reduce the gap between 2D and 3D spaces. \cite{hope} proposes a UNet-based GCNs to estimate the 3D hand pose and the 6D object pose. It learns global geometric relationship constraints for all hand poses. This paper introduces a GCN-based framework for 3D hand pose estimation using a single RGB image.  It exploits the spatial dependencies information between joints to learn per-pose relationship constraints yielding better performance than the global one. The proposed approach does not require depth nor meshes supervision. 

\section{Methodology}
\label{methodology}

Due to the absence of depth information, it is challenging to estimate accurate 3D hand poses from a monocular RGB image. We observe that one of the main obstacles for learning-based 3D hand pose estimation approaches is the non-exploitation of spatial dependencies between adjacent joints to constrain the 3D pose estimation task and reduce the depth ambiguity. Furthermore, \cite{cai2, hope} show that coarse-to-fine techniques improve the 3D pose estimation accuracy. Motivated by these observations, we introduce an effective approach for 3D hand pose estimation from a monocular RGB image (Fig.~\ref{fig:architecture}). 

Our approach is based on GCNs (Section.~\ref{graphrevise}) and includes six main modules. A modified ResNet-10 described in Section.~\ref{featuremaps}. 2D and 3D joint classifiers presented in Section.~\ref{classifiers}. Section.~\ref{relations} explains the relations algorithm that constructs per-pose geometric constraints. This knowledge is combined with a global learnable constraint to regress 2D/3D hand pose (Section.~\ref{regressor}). Section.~\ref{refine} shows our proposed refinement network that employs the ANN algorithm to identify joint relationships improving the estimated pose. Finally, Section.~\ref{loss} presents the loss functions. Note that the architecture of our last five modules is based on GCNs, which are more efficient than CNNs \cite{hope}.

\subsection{Revisiting graph convolution network}
\label{graphrevise}
GCNs are introduced in \cite{graph} to perform semi-supervised classification on graph-structured data. Unlike CNNs that can only handle fixed structures, GCNs treat irregular structures in a non-euclidian space. Let $G=(V, E)$ denote a graph, where $V$ and $E$ represent a set of $M$ nodes and $L$ edges, respectively. Let $A \in [0,1]^{M \times M}$ be the adjacency matrix of $G$. Let  $I$ be the identity matrix and $\tilde{A}=A+I$ is the new self-loop adjacency matrix. Since the graph may include low/high-degree nodes a normalization process is required to avoid vanishing/exploding gradients. \cite{graph} addressed this problem by scaling rows and columns of $\tilde{A} $ in Eq.~\ref{eq:graph_classification2}. 

\begin{equation}
\label{eq:graph_classification2}
 \hat{A} = \tilde{D}^{-\frac{1}{2}} \tilde{A} \tilde{D}^{-\frac{1}{2}}
\end{equation}
Where  $\tilde{D}$ is the degree matrix of $\tilde{A}$. 
GCNs propagate information between nodes to update their representations. The adjacency matrix serves as a mask to select the aggregated nodes. The propagation rule in GCN is computed in Eq.~\ref{eq:graph_classification1}:
\begin{equation}
\label{eq:graph_classification1}
H^{k+1} = \delta ( \hat{A}H^{k}W^{k})
\end{equation}
Where: $\delta=ReLU$ is the activation function, $H^{k}$ and $W^{k}$ are the node features and the weights in the $k^{th}$ layer, respectively. 

\begin{table}[H]
\caption{The architecture of the modified ResNet-10 network}
\begin{tabular}{|c|c|c|c|c|c|}  
\hline
Layer & In & Out & Kernel & Stride & Padding \\ 
\hline
\hline
\centering
Conv2D & 3 & 32 & $7\times7$ & 2 & 3 \\
BN & 32 & 32 & - & - & - \\
ReLU & 32 & 32 & - & - & - \\
Max-pool & 32 & 32 & 3 & 2 & 1 \\
\hline
Res-Block & $32$ & $32$ & $3\times3$ & $1$ & $1$ \\
\hline
Res-Block & $32$ & $64$ & $3\times3$ & $2$ & $1$ \\
Conv2D & $32$ & $64$ & $1\times1$ & $2$ & $0$ \\
BN & $64$ & $64$ & - & - & - \\
\hline
Res-Block & $64$ & $128$ & $3\times3$ & $2$ & $1$ \\
Conv2D & $64$ & $128$ & $1\times1$ & $2$ & $0$ \\
BN & $128$ & $128$ & - & - & - \\
\hline
Res-Block & $128$ & $256$ & $3\times3$ & $2$ & $1$ \\
Conv2D & $128$ & $256$ & $1\times1$ & $2$ & $0$ \\
BN & $256$ & $256$ & - & - & - \\
\hline
Conv2D & $256$ & $21$ & $3\times3$ & $1$ & $1$ \\
\hline
\end{tabular}
\label{table:resnet10}
\end{table}

\subsection{Feature maps extraction}
\label{featuremaps}

We use a modified ResNet-10 architecture \cite{resnet} as a backbone for feature extraction from an RGB hand image. We reduce the number of feature maps in each convolution layer (width) to improve efficiency without affecting the performances. We describe the details of our architecture in Table.~\ref{table:resnet10}. We feed the cropped hand into the network, and we flatten the last two dimensions of the feature maps from $(B, 21, W, H)$ to $(B, 21, W\times H)$, where $B$ is the batch size, $W$ and $H$ are the width and the height of the feature maps, respectively. The reason for this transformation is to get the correct input format for the 2D/3D GCN classifiers, where each joint has a feature vector of size $W\times H$.

\subsection{Classification modules}
\label{classifiers}
We feed the output feature maps of the ResNet-10 network to 2D/3D GCN-based classifiers \cite{graph} to assign each 2D/3D joint to a specific 2D/3D block, respectively, based on its locality. This task is less challenging than regression since the output dimension is restricted. Furthermore, it exploits spatial dependencies between the joints (nodes) to learn 2D/3D per-pose relationship constraints that serve as additional information during pose estimation.

\begin{figure}
\centering
\includegraphics[width=0.45\textwidth]{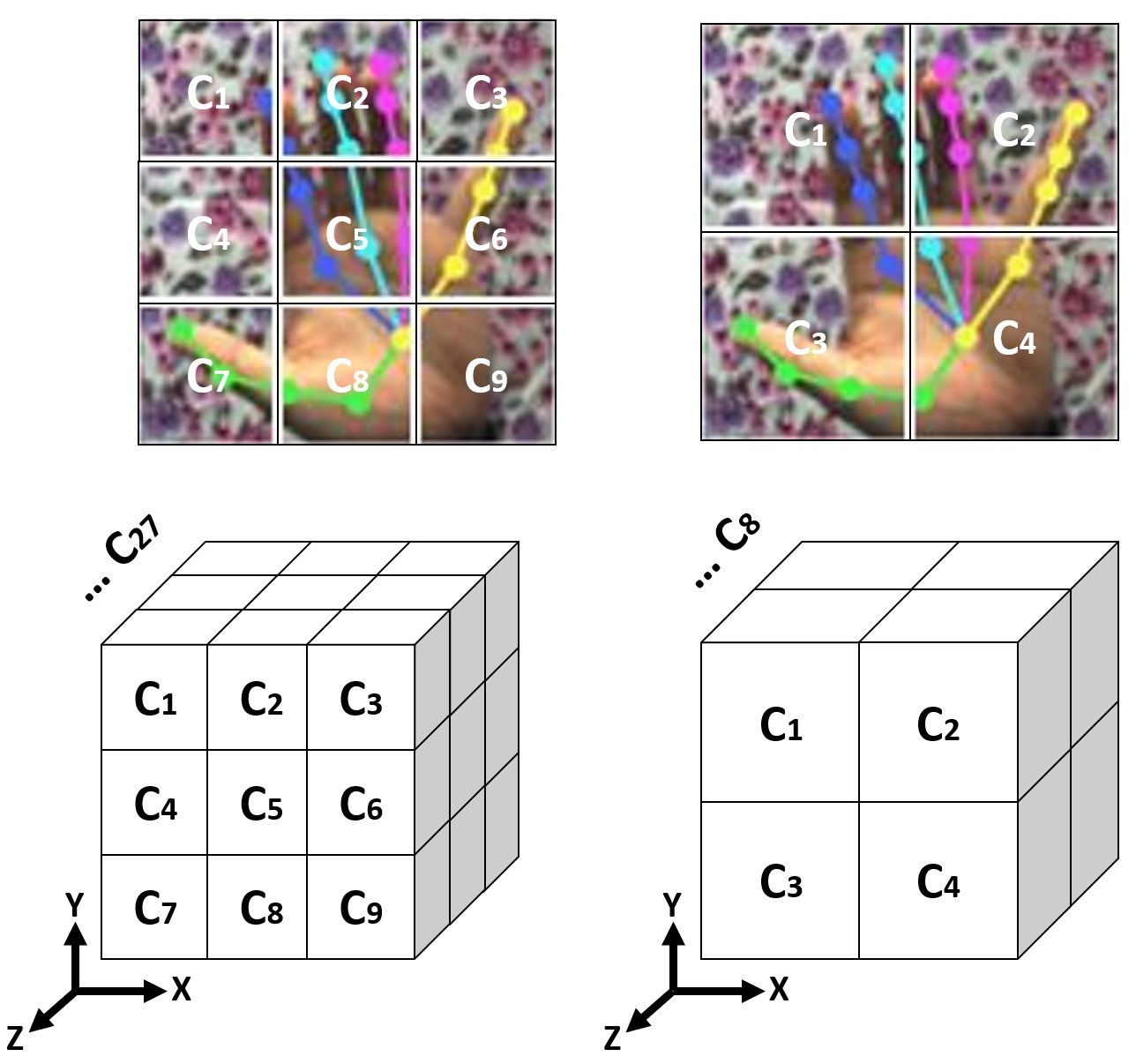}
\caption{We classify each joint position into a predefined number of classes by dividing the 2D and 3D space into discrete intervals in the [$X$, $Y$] and [$X$, $Y$, $Z$] axes, respectively. The top and the bottom rows demonstrate two examples of 2D and 3D quantization. Due to visualization difficulties, we represent the 3D classes by cubes (This figure should be printed in color).}
\label{fig:classification_model}
\vspace{-3mm}
\end{figure}

To train our classifiers, we need the ground truth classes for 2D and 3D joints. Following Algorithm.~\ref{alg:labels2d}, we classify each 2D joint into one of the 2D blocks based on its $(x, y)$ coordinates. Specifically, we divide the 2D space into $N=n \times n$ blocks where $n$ is the number of splits in the horizontal and the vertical axis of the image. The shape of 2D classes is $(B, N, 21)$, where $B$ is the batch size and $21$ is the number of joints. Fig.~\ref{fig:classification_model} illustrates the quantization process of the joints in 2D and 3D spaces. 

In the 3D joint classification, we divide the 3D space into $K=k \times k \times k$ blocks where $k$ is the number of splits in the $x$, $y$ and $z$ axis. Following Algorithm.~\ref{alg:labels3d}, we classify each 3D joint into one of these 3D blocks based on its $(x, y, z)$ coordinates. The shape of 3D classes is $(B, K, 21)$ and the output of the 2D/3D classifier in Fig.~\ref{fig:architecture} is a tensor of class probabilities for each joint.

In this stage, we use the propagation rule in \cite{hope} to classify 2D/3D joints. Specifically, our GCNs learn global adjacency matrix $\hat{A}$ (constraints) instead of feeding a predefined one \cite{graph}. We empirically set the number of graph convolution layers of our network to two, where each layer is defined in Eq.~\ref{eq:graph_classification1}. We feed the estimated values to the \textit{Relations} function in Algorithm.~\ref{alg:relations}, which outputs a per-pose 2D/3D geometric relationship constraints to guide the pose estimation task. In particular, 2D/3D per-pose adjacency matrix expresses spatial dependencies between each pair of joints. We find that the GCN-based classifier exhibits better performance than Fully Connected (FC) networks. Note that we quantized the 2D/3D interval into overlapped blocks to consider the problem of joints residing in borders. However, we do not notice any gain over the proposed classification approach. We can explain this by that using overlapped blocks increases the number of classes reducing accuracy performances. Also, a joint can belong to different classes, which confuses the classifier.

\subsection{Relations function}
\label{relations}

We employ the \textit{Relations Function} described in Algorithm.~\ref{alg:relations} to express geometric constraints between the joints. It constructs a 2D/3D per-pose adjacency matrix that serves as an extra weighting for the GC layers. The algorithm takes the output of the 2D/3D classifier, applies a \textit{Softmax} activation function to get class probabilities of each joint and an \textit{Argmax} function to find the class corresponding to the highest probability. Finally, adjacent joints are identified by comparing their labels. The output matrix is of size $21 \times 21$ where each $(i,j)$ coordinate exhibits the relation between $joint_i$ and $joint_j$. Note that the proposed function can identiafy relations between pairs, which are physically disconnected in the hand skeletal model. 

\begin{algorithm}
\SetKwInOut{Input}{Input}
\SetKwInOut{Output}{Output}
\underline{function RelationsFunction} $(data)$\;
\Input{Classifier outputs for each joint.}
\Output{$21 \times 21$ relations matrix.}
$matrix \leftarrow []$\;
$probabilities \leftarrow softmax(data, dim=1)$\;
$labels \leftarrow argmax(probabilities, dim=1)$\;
\For{$i$=$0$:$21$}
{
$values \leftarrow []$\;
\For{$j$=$0$:$21$}
{
$a \leftarrow labels[:, i]$\;
$b \leftarrow labels[:, j]$\;
$values[j] \leftarrow (a == b)$\;
}
$matrix[i] \leftarrow values$\;
}
return $matrix$\;
\caption{Relations function.}
\label{alg:relations}
\end{algorithm}

\subsection{Regression modules}
\label{regressor}

This module aims to estimate 21 2D/3D hand joints from a monocular RGB image. \cite{hope} estimates the 3D poses using Eq.~\ref{eq:graph_classification1} as a propagation rule. It learns a global adjacency matrix between all the poses in the dataset. Although \cite{hope} outperforms the limitation of \cite{cai2, ge} that do not include relations of physically disconnected joints, it presents another problem. In particular, learning only a global adjacency matrix is a potential limitation since different hand poses in the dataset share a single relationship constraint (e.g. open and close hands have the same adjacency matrix despite being physically dissimilar). Our 2D/3D GCN-based regressors exploits the estimated 2D/3D per-pose adjacency matrices $R_{2D}$ and $R_{3D}$ as local constraints for each hand pose (Eq.~\ref{eq:graph_regression2d_1} and  Eq.~\ref{eq:graph_regression3d_rel3d}). Besides, It keeps the learnable adjacency matrix $\hat{A}$ \cite{hope} that expresses global relationships. This combination guides the 2D/3D hand pose regressors to provide an initial promising pose using Eq.~\ref{eq:graph_regression2d_2} and Eq.~\ref{eq:graphregression_3d}. 
 
\begin{equation}
\label{eq:graph_regression2d_1}
  R_{2D} = RelationsFunction(L_{2D})
\end{equation}

\begin{equation}
\label{eq:graph_regression2d_2}
  H_{2D}^{k+1} = ReLU ([\hat{A}H_{2D}^{k}W_{A}^{k}, R_{2D}H_{2D}^{k}W_{R}^{k}])
\end{equation}

Where: $L_{2D}$ is the output of the 2D classifier, $H_{2D}^{0}$ is the concatenation of the ResNet-10 features and the 2D classifier probabilities. $W_{A}^{k}$ and $W_{R}^{k}$ are the weights for the global and the local adjacency matrices, respectively.

Recent works demonstrate that 2D pose information improves the 3D pose estimation task \cite{semantic, simple, fang}. We feed the concatenation of the estimated 2D poses, the image feature maps of the ResNet-10 network, the 2D classification probabilities and the 3D classification probabilities ($H_{3D}^{0}$) to the 3D regressor. Furthermore, we exploit the 3D relations matrix (Eq.~\ref{eq:graph_regression3d_rel3d}) as an additional weighting mechanism during neighborhood aggregation to express the spatial dependencies between the 3D joints (Eq.~\ref{eq:graphregression_3d}).

\begin{equation}
\label{eq:graph_regression3d_rel3d}
  R_{3D} = RelationsFunction(L_{3D})
\end{equation}

\begin{equation}
\label{eq:graphregression_3d}
  H_{3D}^{k+1} = ReLU ( [\hat{A}H_{3D}^{k}W_{A}^{k}, R_{3D}H_{3D}^{k}W_{R}^{k}])
\end{equation}

Where: $L_{3D}$ is the output of the 3D classifier.

\subsection{3D refinement module}
\label{refine}

After getting reliable 2D/3D poses from the coarse stage, we freeze the network weights, and we improve the estimated 3D pose through a refinement module. The latter applies the proposed ANN algorithm to construct a per-pose adjacency matrix for the GCN. Previous works employ KNN-based graphs for other computer vision tasks. \cite{link,knowledge} create graphs from image features and apply residual KNN-based GCN for face clustering. \cite{seg} connects KNN 3D points cloud for RGBD semantic segmentation. However, fixing a unique number of nearest neighbors $K$ for all the hand poses of the dataset does not capture the underlying structure of joints, which may affect information aggregation in the GCN.  Moreover, the same hand pose may include high-degree nodes (joints) as well as low-degree ones. Hence, the number of nearest neighbors should be different for each joint based on its distance to the others. 

To alleviate this problem, we first calculate the distance matrix $D \in R^{21 \times 21}$ among $21$ joints. Subsequently, we learn a global threshold $T$ to be the criteria to select the related joints. Following Algorithm.~\ref{alg:ann}, if $D(i, j) \leq T$ then joint $i$ and joint $j$ are considered as adjacent and $R_{ANN}(i, j) = 1$, where $R_{ANN}$ is the adjacency matrix. We feed  $R_{ANN}$ and the prior estimated pose $H_{3D}^{0}$ to GCN refinement to produce a more accurate pose (Eq.~\ref{eq:graphregression_3d_refine}). Note that the gain of the adaptative ANN-based GCN module appears when the coarse pose is predicted from our hybrid classification-regression approach. In other words, estimating 3D poses without using the proposed local constraints (2D/3D per-pose adjacency matrix) produces less accurate coarse poses. Thus, ANN may identify wrong nearest neighbors in the adjacency matrix.

\begin{equation}
\label{eq:graphregression_3d_refine}
  H_{3D}^{k+1} = ReLU ( [\hat{A}H_{3D}^{k}W_{A}^{k}, R_{ANN}H_{3D}^{k}W_{R}^{k}])
\end{equation}

\begin{algorithm}
\SetKwInOut{Input}{Input}
\SetKwInOut{Output}{Output}
\underline{function ANN} $(3D, \theta)$\;
\Input{$3D$: coarse predictions, $\theta$: learnable threshold}
\Output{$21 \times 21$ ANN relations matrix.}
$matrix \leftarrow []$\;
\For{$i$=$0$:$21$}
{
$values \leftarrow []$\;
\For{$j$=$0$:$21$}
{
$values[j] \leftarrow MSE(3D[i],3D[j])$\;
}
$matrix[i] \leftarrow values$\;
}
$matrix = float(matrix <= \theta)$\;
return $matrix$\;
\caption{Adaptative nearest neighbour algorithm.}
\label{alg:ann}
\end{algorithm}

\subsection{Loss functions}
\label{loss}

Our training schema is composed of two stages coarse and refinement. The former involves two loss terms for the classification and the regression. Eq.~\ref{eq:classification2d} and Eq.~\ref{eq:classification3d} classify each joint into one of the blocks in the 2D/3D space, where $CE(.)$ is a multi-class Cross-entropy loss function. In Eq.~\ref{eq:regression2d} and Eq.~\ref{eq:regression3d}, we use an $L2$ regression loss between the predicted and the ground truth 2D/3D joints.  

\begin{equation}
\label{eq:classification2d}
  L_{classification2d} = \sum_{i=1}^{21} CE_i(PredictedClass2D_i, GTClass2D_i)
\end{equation}

\begin{equation}
\label{eq:classification3d}
  L_{classification3d} = \sum_{i=1}^{21} CE_i(PredictedClass3D_i, GTClass3D_i)
\end{equation}

\begin{equation}
\label{eq:regression2d}
  L_{regression2d} = L_2(Predicted2DPose, GT2DPose)
\end{equation}

\begin{equation}
\label{eq:regression3d}
  L_{regression3d} = L_2(Predicted3DPose, GT3DPose)
\end{equation}

The coarse loss in Eq.~\ref{eq:total} is a linear combination of the regression and the classification loss, where $\delta_1=100$ and $\delta_2=1$ for the regression and the classification terms, respectively. We empirically find the best hyperparameters by trying different configurations. 

\begin{equation}
\label{eq:total}
  L_{coarse} = \delta_1 * (L_{regression2d}+L_{regression3d}) + \delta_2 * (L_{classification2d}+L_{classification3d})
\end{equation}

In the refinement stage, we use the $L2$ loss between the ground truth $3D$ and the predicted one (Eq.\ref{eq:refinement}).

\begin{equation}
\label{eq:refinement}
  L_{refimenent} = L_2(Fine3DPose, GT3DPose)
\end{equation}

\section{Experiments}
\label{experimental}

\subsection{Implementation details}
\label{implementation}

We implement the proposed method using Pytorch v1.6 \cite{pytorch}, CUDA v10.1 and cuDNN v7.6.4. We train our model on images of size $256\times256$ for 400 epochs using a batch size of 64. We initialize the weights of the ResNet-10 network using a normal distribution, where the $mean=0$ and $std=0.02$. We initialize the weights of the GCN networks using Xavier \cite{xavier} where we set the gain to the square root of $2$. We use NVIDIA's Apex mixed-precision training of 16-bits to speed-up the training. It takes three days to converge on a single NVIDIA TITAN X GPU.  We empirically found that AdaDelta optimizer \cite{adadelta} is better than ADAM \cite{adam} for our task. Since the weights updating formula does not include a learning rate, we use the default optimizer parameters.

\subsection{Datasets and evaluation metrics}
\label{datasets}
To train and evaluate our proposed method, we need datasets that include RGB images with their 2D and 3D annotations. We conduct our experiments using the Stereo Hand Pose Tracking Benchmark (STB) \cite{stb} [dataset] and Rendered Hand Pose (RHD) [dataset] datasets \cite{zimmermann}. These benchmarks are widely used for 3D hand pose estimation from a monocular RGB image.

STB \cite{stb} contains 30k and 6K real-world images for training and test, respectively. This dataset provides the RGB images with their corresponding 3D annotations. To obtain the 2D keypoints, we perform a projection operation from the 3D space using the given camera intrinsic parameters. This dataset is less challenging since it has less lighting, viewpoint, and background variations. Also, the images are not noisy and have high resolution ($640\times480$).

RHD \cite{zimmermann} is a very challenging hand dataset that contains synthetic images of resolution $320\times320$ collected from twenty persons performing 39 actions. The images are rendered from several viewpoints and using different illumination conditions. RHD dataset contains 41K images for training and 3K images for testing covering several backgrounds and hand shapes. Each sample in the dataset includes an RGB image, a hand mask image, the corresponding depth image, the 2D key-points and the 3D hand pose.

To crop the hand from the image, we find the $min_x$, $min_y$, $max_x$ and $max_y$ of the 2D points. The top left corner $(min_x, min_y)$ and the bottom right corner $(max_x, max_y)$ are subtracted/added to a small threshold of 10 and 20 for the RHD and STB datasets, respectively, ensuring that the cropped hand image includes all the joints.

To evaluate our approach and quantitively compare it against the state-of-the-art methods, we report the common metric mean end-point-error (EPE), which measures the average Euclidean distance between the ground-truth and the estimated keypoint. The distances are expressed in millimeters (mm) and pixels (px) for 2D and 3D hand pose estimation, respectively. Furthermore, we show the area under the curve (AUC) on the percentage of correct keypoints (PCK), which is widely used to evaluate human/hand pose estimation approaches with different thresholds. Although the main aim of the proposed method is to estimate the 3D hand pose, we also validate the classification performance using common metrics, including accuracy, precision and recall.

\begin{figure}
\centering
\includegraphics[width=0.4\textwidth]{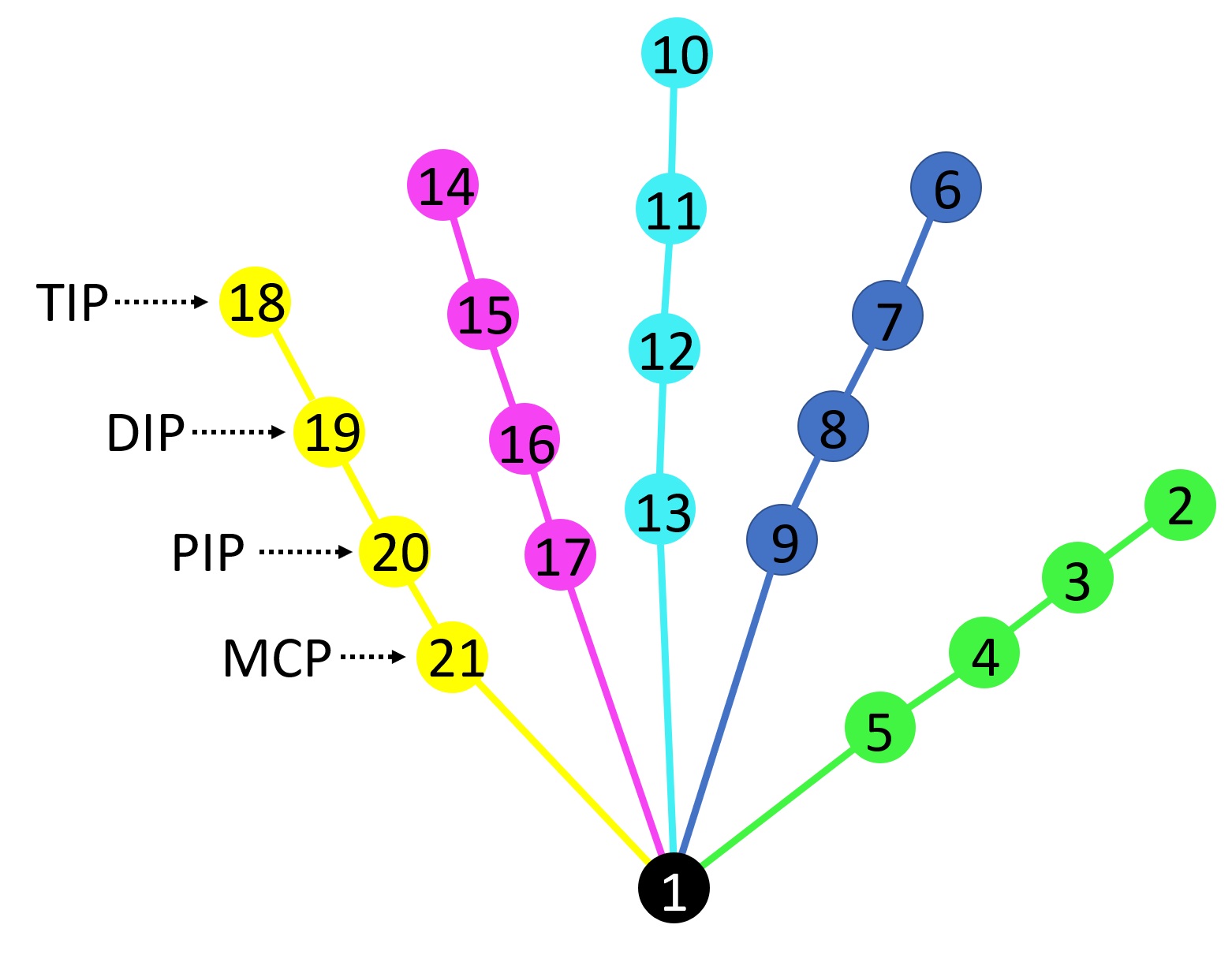}
\caption{The description of our hand model, where the first joint is the wrist. For each finger from the thumb to the pinky, the order starts from the tip to the mcp (This figure should be printed in color).}
\label{fig:hand_model}
\vspace{-3mm}
\end{figure}

\subsection{Hand model representation}
\label{hand-model}

We follow the hand model of \cite{zimmermann, stb} (Fig.~\ref{fig:hand_model}), such that the hand is described by 21 joints, namely the wrist representing the hand palm and four joints (tip, dip, pip and mcp) per finger (thumb, index, middle, ring and pinky). Since we aim to identify the spatial relationships between each pair of joints in 2D/3D space, the order of the joints and fingers is important to understand the proposed per-pose constraints (adjacency matrices). We define each joint position by $joint_{2D}=(x, y)$ and $joint_{3D}=(x, y, z)$ for 2D and 3D coordinates, respectively. 

\subsection{Self-comparisons}
\label{results}

To evaluate the proposed approach and find the best design choice, we conduct extensive experiments on the RHD dataset since it is more challenging. We first investigate the impact of our hybrid classification-regression approach (per-pose constraints) on the hand pose estimation performances. We analyze the effect of the number of classes, and we show the benefit of the proposed ANN-based refinement module. We select the best model to run it on the STB dataset and compare it against the state-of-the-art methods.

\subsubsection{Analysis of the classification modules (per-pose constraints)}
\label{effect-class}

To validate the proposed hybrid classification-regression method, we perform two different experiments. The first one is \textit{Baseline A} that employs ResNet-10 and 2D/3D GCN-based regressors without the refinement module. Furthermore, it uses only the global learned relationship constraints (adjacency matrix) to estimate the 2D/3D hand pose, respectively. In particular, we remove the classification modules that provide 2D/3D per-pose constraints from the framework in Fig.~\ref{fig:architecture}. In the second experiment, we implement the proposed approach (\textit{Full}) that classifies the joints into several 2D/3D blocks to find a spatial relationship between the joints. We employ this information as an individual adjacency matrix for each sample of the dataset (per-pose constraint) to estimate promising coarse poses. In the second stage, the ANN-based GCN refines the initial 3D pose to generate a more accurate one. We can see from Table.~\ref{table:self} that our proposed method (\textit{Full}) outperforms \textit{Baseline A} that provides lower scores in both AUC and EPE metrics proving the superiority of the proposed per-pose constraint over the global shared one. 

\begin{table}
\caption{Ablation studies of different baselines with mean EPE [mm] and AUC on the RHD dataset (for EPE lower is better, for AUC higher is better). }
\label{table:self}       
\begin{tabular}{|c|c|c|}
\hline
Method  &$EPE^-$  & $AUC^+$   \\
\hline
\hline
Baseline A: full w/o classification & 19.16 & 0.858  \\  
\hline
Baseline B: full w/o refinement (coarse) & 14.76 & 0.927  \\ 
\hline
Baseline C: full w/ FC refinement & 14.68 & 0.928  \\ 
\hline
Baseline D: full w/ KNN refinement & 13.43 & 0.935  \\ 
\hline
Full  &\textbf{12.81} &\textbf{0.939}  \\ 
\hline
\end{tabular}
\end{table}

\subsubsection{Effect of the number of classes}
\label{effect-classnumber}

We conduct a set of experiments to find the best number of 2D/3D blocks (classes) that provide the best performances. We compare the effect of 4, 9, 16 and 25 classes by dividing the 2D space into ($2\times2$), ($3\times3$), ($4\times4$) and ($5\times5$) blocks, respectively. Besides, we compare the effect of 8, 27, 64 and 125 classes by dividing the 3D space into ($2\times2\times2$), ($3\times3\times3$), ($4\times4\times4$) and ($5\times5\times5$) blocks, respectively. Table.~\ref{table:classification} shows that our GCN-based classifiers yield high performance if the number of classes is less than 27. We note that having a large number of classes produces an identity-like adjacency matrix. Meanwhile, having a small number of classes increases the adjacent joints, and the matrix will not effectively express relations. We empirically fix the best number of classes to 16 and 27 for the 2D and 3D classifiers, respectively. This configuration achieves high classification performances and constructs meaningful per-pose relation matrices improving the 2D/3D pose estimation. 

\begin{table}
\caption{Effect of the number of classes on our 2D/3D classifiers. Each triplet in the classification metrics represents the accuracy, recall and precision.}
\label{table:classification}       
\begin{tabular}{|c|c|c|c|}
\hline
2D classes & 2D classification metrics & 3D classes & 3D classification metrics  \\
\hline
\hline
4 & 99.98/99.94/99.88 & 8 & 99.97/99.92/99.82 \\ 
\hline
9 & 99.96/99.90/99.81 & 27 & 99.83/99.78/99.71 \\ 
\hline
16 & 99.91/99.87/99.78 & 64 & 95.15/94.92/94.53 \\ 
\hline
25 & 99.86/99.82/99.73 & 125 & 88.11/86.74/85.23 \\ 
\hline
\end{tabular}
\end{table}

\begin{figure*}
\centering
\includegraphics[width=0.99\textwidth, height=0.41\textheight]{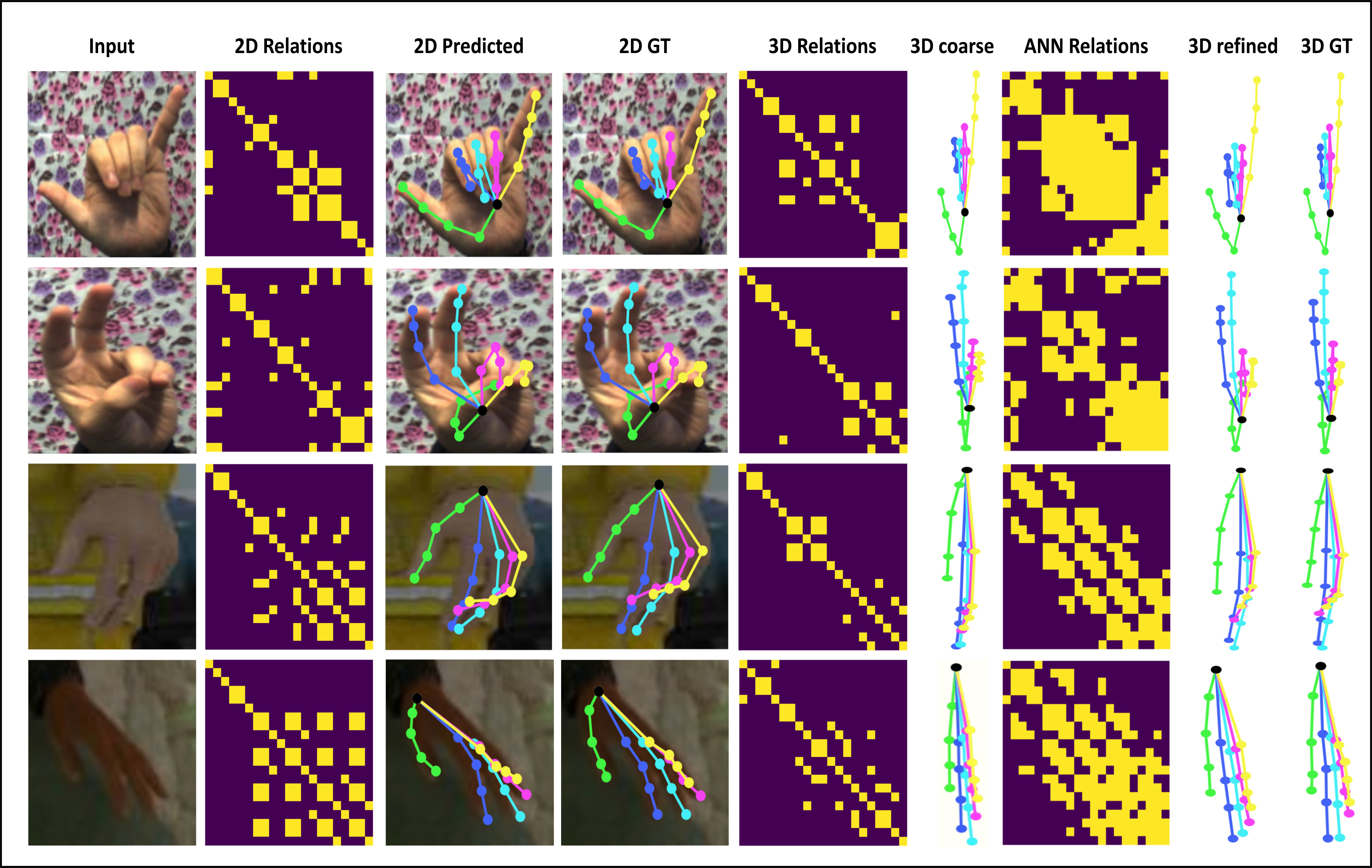}
\caption{The intermediate output of the proposed method on STB and RHD datasets. From left to right, we show the input hand image, the per-pose 2D relations matrix, the predicted 2D key-points, the ground truth 2D keypoints, the per-pose 3D relations matrix, the predicted coarse 3D pose, the per-pose ANN relations matrix, the refined 3D pose and the ground truth 3D pose (This figure should be printed in color).}
\label{datasets_relations}
\vspace{-3mm}
\end{figure*}

\subsubsection{Effect of the refinement network }
\label{effect-refine}

As mentioned previously, our approach estimates an initial 3D hand pose in the coarse stage. In the second stage, the refinement network takes as inputs the estimated coarse pose and the learned ANN adjacency matrix to obtain a more accurate pose. To investigate the impact of this module, we report the quantitative results of the coarse prediction (\textit{Baseline B}) in Table.~\ref{table:self}. Also, Fig.~\ref{datasets_relations} shows the qualitative results of the coarse estimation, the refined estimation and the learned per-pose adjacency matrices. We can see that the refinement module improves the coarse prediction and achieves more reliable qualitative and quantitative results. We can explain this by that the proposed ANN technique uses the estimated coarse 3D poses to provide a precise per-pose adjacency matrix. Hence, exploiting this information at this stage yields more accurate spatial dependencies and improves performance. Table.~\ref{table:self} shows a significant margin between the full model and \textit{Baseline A}. The latter estimates the coarse pose without exploiting per-pose constraints to identify spatial relations. Therefore, applying the proposed ANN on \textit{Baseline A} identifies wrong related joints and lead to poor performance.

The GCN-based refinement modules outperform the simple FC refinement module (\textit{Baseline C}) demonstrating the superiority of GCN to fit the problem of hand pose estimation. To validate the proposed ANN algorithm, we compare it against KNN (\textit{Baseline D}), where we empirically fix K to 5. Results show that the ANN-based refinement module outperforms the KNN-based version (Table.~\ref{table:self}). We can explain this by that the KNN adjacency matrix learns the same number of adjacent joints regardless of the pose and the joint locations. On the other hand, we can see from (Fig.~\ref{datasets_relations}) that the number of neighboring joints in the ANN adjacency matrix is based on the joints distribution in the pose. We note that after training, the model learns the threshold parameter $T_{STB}=0.052$ and $T_{RHD}=0.0448$ for the two datasets.

\subsection{Qualitative results}
\label{qualitative}

In addition to the quantitative validation, we show qualitative results of the proposed 2D/3D hand pose estimation method on STB and RHD datasets. Fig.~\ref{datasets_relations} presents all the intermediate outputs, including the predicted poses and the learned per-pose adjacency matrices. We can see that the estimated 2D and 3D poses are very close to the ground truth, demonstrating that the proposed approach achieves accurate results for both 2D and 3D hand pose estimation. Furthermore, the 2D, 3D and ANN adjacency matrices reflect spatial relationships between the predicted joints that can be physically connected or disconnected in the hand skeletal.

\begin{figure*}
\centering
\includegraphics[width=0.99\textwidth, height=0.45\textheight]{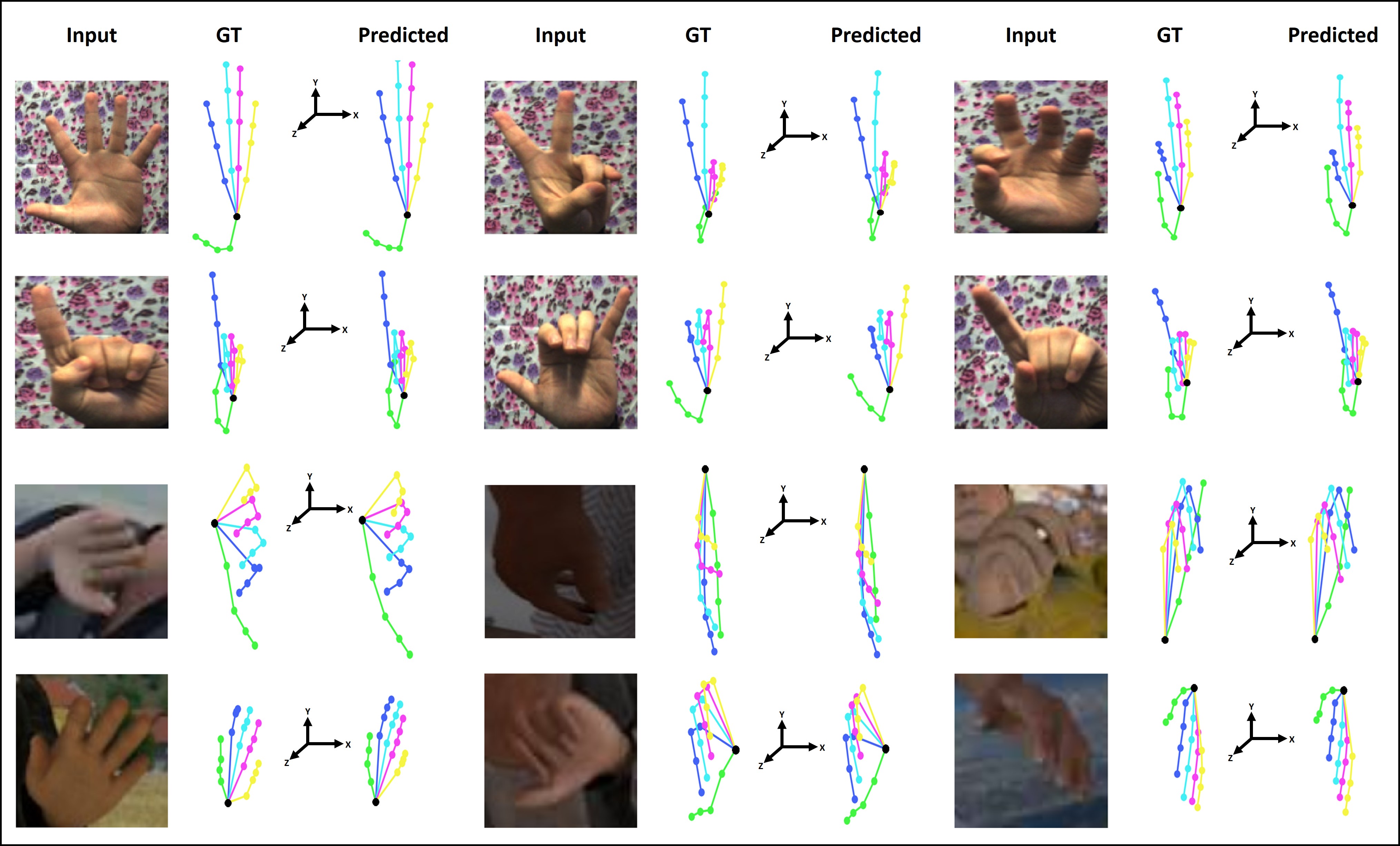}
\caption{ 3D hand pose estimation results on STB (first two rows) and RHD (last two rows) datasets. For each triplet, the left column represents the input RGB image, the middle column is the ground truth 3D joint skeleton and the right column is our corresponding prediction (This figure should be printed in color).}
\label{3d_results}
\vspace{-3mm}
\end{figure*}

Fig.~\ref{3d_results} shows more qualitative results of the 3D hand pose estimation task on STB and RHD datasets. We can see that the estimated 3D poses are reliable even in case of self-occlusion and complex background textures. We also report more qualitative results for 2D hand pose estimation in Fig.~\ref{2d_results}.

\begin{figure}
\centering
\includegraphics[width=0.49\textwidth]{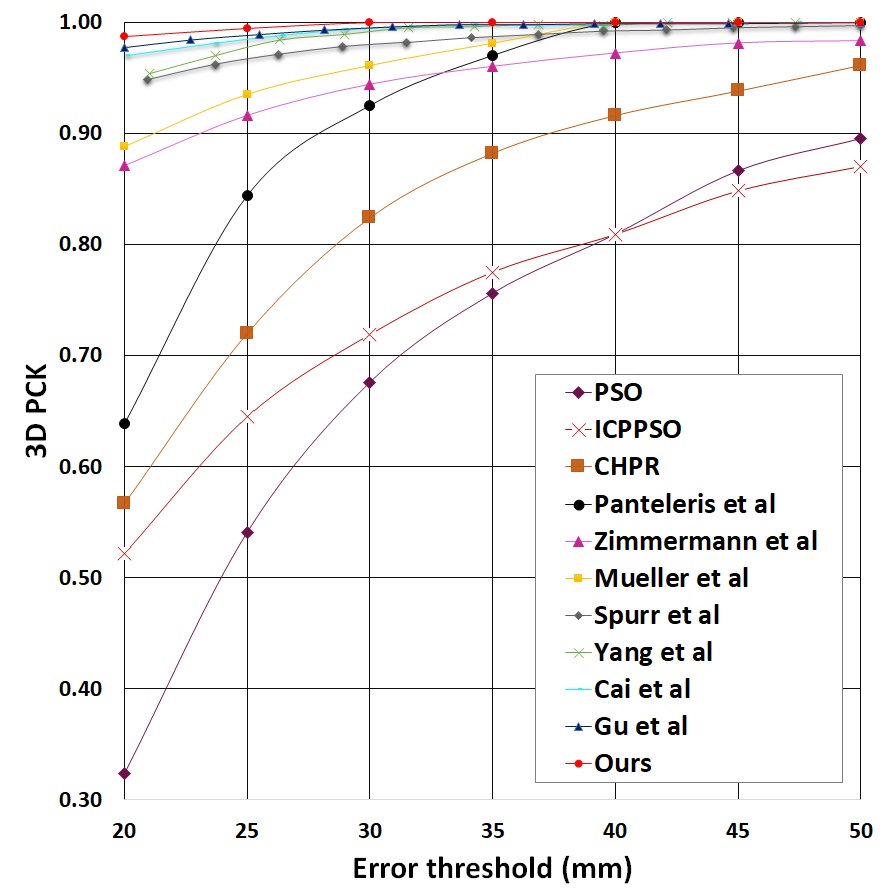}
\caption{Comparison with the state-of-the-art methods \cite{pso, chpr, icppso, panteleris, zimmermann, mueller, spurr, yang, cai2, gu} on the STB dataset using 3D PCK (This figure should be printed in color).}
\label{stb_pck}
\vspace{-3mm}
\end{figure}

\begin{figure}
\centering
\includegraphics[width=0.49\textwidth]{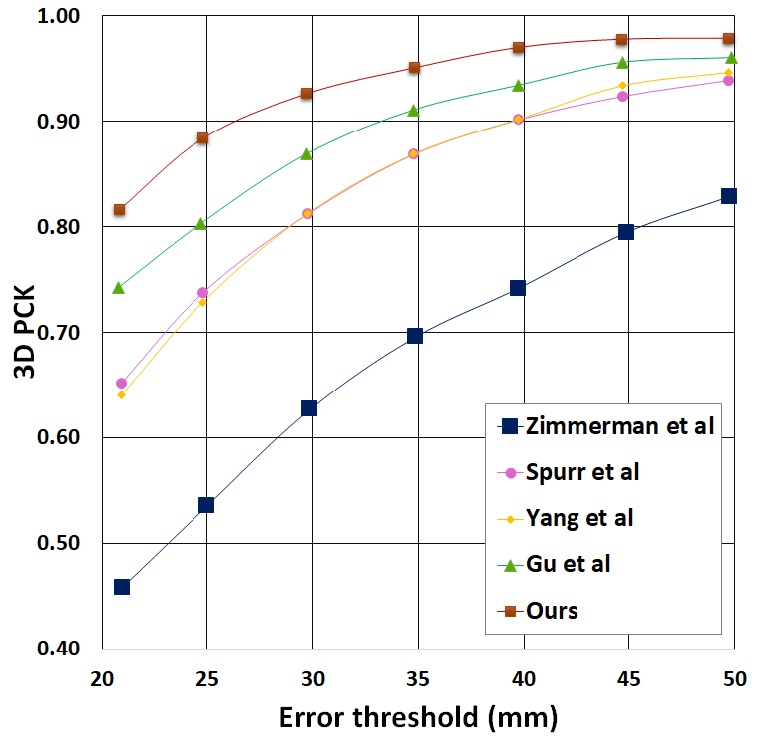}
\caption{Comparison with the state-of-the-art methods \cite{zimmermann, spurr, yang, gu} on the RHD dataset using 3D PCK (This figure should be printed in color).}
\label{rhd_pck}
\vspace{-3mm}
\end{figure}

\begin{table}
\centering
\caption{Comparison with the state-of-the-art methods on RHD and STB datasets using Mean EPE [pixel].  
}
\begin{tabular}{|c|c|c|}
\hline
Methods & RHD & STB\\
\hline
Zimmermann et al\cite{zimmermann} &9.14  & 5 \\
\hline
Iqbal et al \cite{iqbal} & 3.57 &  -\\
\hline
Ours  &\textbf{2.91} & \textbf{1.56}\\
\hline
\end{tabular}
\label{2D}
\end{table}

\begin{table}
\centering
\caption{Comparison with the state-of-the-art methods on RHD and STB datasets using AUC.   }
\begin{tabular}{|c|c|c|}
\hline
Methods & RHD & STB\\
\hline
\hline
Zimmermann et al \cite{zimmermann} & 0.675 & 0.948 \\
\hline
Mueller et al \cite{mueller} & - & 0.965 \\
\hline
Spurr et al \cite{spurr}& 0.849& 0.983 \\
\hline
Cai et al \cite{cai}&0.887 & 0.994 \\
\hline
Iqbal et al \cite{iqbal} & - &0.994  \\
\hline
Yang et al \cite{yang}& 0.849 & 0.991 \\
\hline
Boukhayma et al \cite{boukhayma} & - &0.994  \\
\hline
Baek et al \cite{baek} & 0.926 &0.995  \\
\hline
 Ge et al \cite{ge}& 0.920 & 0.998 \\
 \hline
Zhang et al \cite{zhang}& 0.901 & 0.995 \\
 \hline
Cai et al \cite{cai2} & - & 0.995 \\
 \hline
Guo et al \cite{guo}& 0.933 & 0.998 \\
\hline
Zhao et al \cite{zhao}& 0.872 &0.987  \\ 
\hline
Zhou et al\cite{zhou}& 0.856 &  0.898\\ 
\hline
Gu et al \cite{gu} & 0.887 &0.996  \\
\hline
Ours  & \textbf{0.939}  & \textbf{0.998} \\
\hline
\end{tabular}
\label{table:auc}
\end{table}

\begin{table}
\centering
\caption{Comparison with the state-of-the-art methods on RHD and STB datasets using Mean EPE [mm].  }
\begin{tabular}{|c|c|c|}
\hline
Methods & RHD & STB\\
\hline
\hline
Zimmermann et al \cite{zimmermann} &30.42  & 8.68 \\
\hline
Spurr et al \cite{spurr}& 19.73 &8.56 \\
\hline
Yang et al \cite{yang}& 19.95 &8.66  \\
\hline
Gu et al \cite{gu} &17.11  & 7.27 \\
\hline
Zhao et al \cite{zhao}& - &8.18  \\ 
\hline
Ours  &\textbf{12.81} & \textbf{6.58}\\
\hline
\end{tabular}
\label{table:epe}
\end{table}

\subsection{Comparisons with state-of-the-art method}
\label{sota}

We report the quantitative results of the 2D hand pose estimation task with EPE (pixel), and we compare it against \cite{zimmermann,iqbal}. We can see from Table.~\ref{2D} that our approach achieves low errors in the two datasets and outperforms the state-of-the-art, proving that the 2D GCN-based regressor benefits from the learned per-pose constraint (2D relations matrix). 

Since our main aim is to estimate the 3D hand pose, we compare the proposed method against several competitive state-of-the-art on RHD and STB datasets with all the aforementioned metrics. For a fair comparison, we follow previous works preprocessing of the RHD dataset \cite{zimmermann}, and we validate the proposed method under the same evaluation conditions and standards and using the same data split. We use a range of $20-50$ (mm) for the 3D PCK metric since it is a well-known standard criterion for RHD and STB dataset. Fig.~\ref{stb_pck} shows the 3D PCK results comparison on the STB dataset of our approach and several state-of-the-art methods \cite{pso, icppso, chpr, panteleris, zimmermann, mueller, spurr, yang, cai2, gu}. Our method outperforms \cite{pso,chpr,icppso, panteleris, zimmermann, mueller, spurr, yang} and achieves competitive performances on the STB dataset to \cite{cai2,gu}. To further evaluate our approach, we report the 3D PCK curves of the RHD dataset. Fig.~\ref{rhd_pck} shows that our method outperforms the compared state-of-the-art methods \cite{zimmermann, spurr, yang, gu} on all the PCK thresholds. 

We report the AUC of 3D PCK for a better understanding of the results. Table.~\ref{table:auc} presents the results of the two datasets with additional state-of-the-art methods \cite{iqbal, cai, ge, boukhayma, zhao, zhou, baek, zhang, guo}. We can see that the proposed approach has the highest score in the RHD dataset, improving the AUC value to 0.939. Also, there is a small improvement in the STB dataset (0.998). Note that the advantage of our approach appears in the RHD dataset that includes low-resolution images with complex backgrounds and poses. STB dataset is less challenging since it contains simple backgrounds with a restricted number of poses. Finally, Table.~\ref{table:epe} lists the EPE metric comparison against five state-of-the-art methods \cite{zimmermann,spurr,yang,gu,zhao}. We can see that the proposed method has superior performances providing 12.81 and 6.58 mean EPE on RHD and STB datasets, respectively. This is mainly due to the learned per-pose geometric constraint (adjacency matrix) and the ANN-based refinement module.

\subsection{Computation complexity}
\label{computation}

Although the proposed framework includes six modules, it is computationally efficient, where the running time on NVIDIA TITAN X GPU  is 7ms. We can explain this by that the network consists of a lightweight feature extractor (ResNet10) and five GCN-based modules, which are significantly faster than convolutional neural networks \cite{hope}. While the classification modules (per-pose constraints) notably contributes to improving the 3D hand pose estimation accuracy, it necessitates only an additional 1ms per image, which does not speed down the inference time.

\section{Conclusions}
\label{conclusion}

In this work, we propose an efficient GCN-based framework for 2D/3D hand pose estimation from a monocular RGB image. Our method addresses one of the key challenges of GCN by learning per-pose geometric constraints expressing local and global relationships between the hand joints. We obtain this information using a classification step that exploits the spatial dependencies between the joint. We improve the predicted pose through a GCN-based refinement module that employs the proposed ANN algorithm to identify joint relationships based and we obtain more accurate 3D hand poses. Experimental results on realistic and synthetic datasets prove the competitiveness of our efficient approach compared to several state-of-the-art methods. In our future work, we plan to exploit temporal information to improve the overall model performance.

\section*{Declaration of Competing Interest}
The authors declare that they have no known competing financial interests or personal relationships that could have appeared to influence the work reported in this paper.

\section*{Acknowledgement}
This research did not receive any specific grant from funding agencies in the public, commercial, or not-for-profit sectors.

\bibliographystyle{cas-model2-names}

\bibliography{cas-refs}
\clearpage

\appendix

\counterwithin{figure}{section}

\section*{Supplementary material}
\section{2D qualitative results}
\label{2D-poses}

\begin{figure*}
\centering
\includegraphics[width=0.99\textwidth]{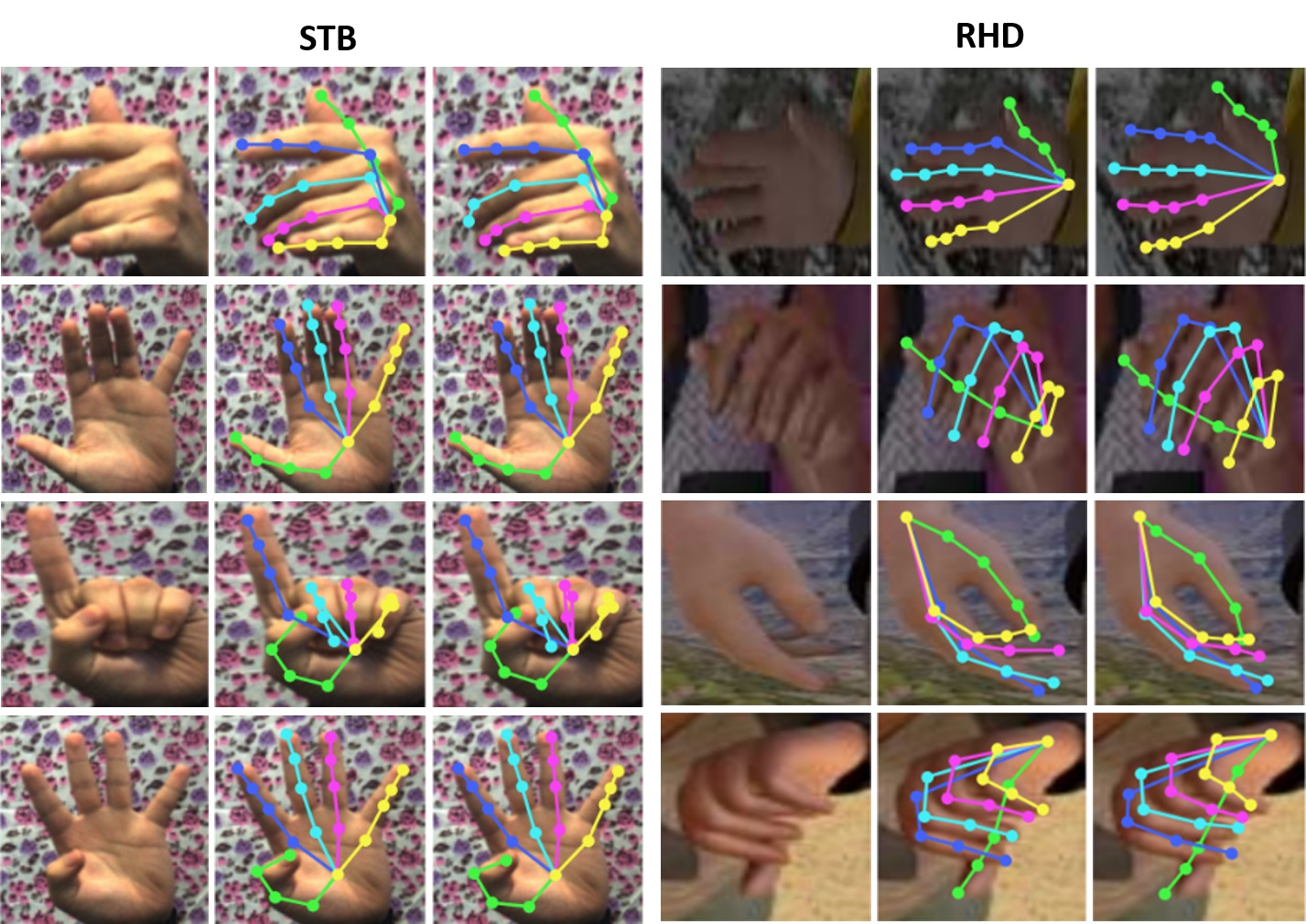}
\caption{ 2D hand pose estimation results on STB and RHD datasets. For each triplet, the left column represents the input RGB image, the middle column is the ground truth 3D joint skeleton and the right column is our corresponding prediction (This figure should be printed in color).}
\label{2d_results}
\vspace{-3mm}
\end{figure*}

We provide qualitative results for 2D hand pose estimation in Fig.~\ref{2d_results}. This task also has several applications, such as sign language and gesture recognition. We can see that the proposed approach produces accurate 2D poses on STB and RHD datasets.

\section{Ground truth of 2D/3D joints classification task}
\label{appendix_gorund truth}

In this section, we explain the creation of the ground truth 2D/3D classes. The function \textit{CreateClasses2D} in Algorithm.~\ref{alg:labels2d} creates an array of size $21$, where each value represent the class of a specific 2D joint. The inputs are the $(x, y)$ coordinates of the 2D keypoints, the number of splits and the image size (the width and height must be equal). Following, we explain the algorithm by specifying the line number and a brief description of its functionality. Note that our algorithms follow a Pythonic pseudo-code style.

\begin{itemize}
\item \textit{Line 2} calculates the size of each block.
\item \textit{Line 6 to 8} initializes the \textit{$parts$} array, which represents the beginning and the end of the blocks in the $x$/$y$ axis, e.g.: if the number of splits is 4, \textit{$parts=[0, 64, 128, 192, 256]$}, where $C_0 \in [0, 64]$, $C_1 \in [64, 128]$, etc.
\item \textit{Line 9 to 14}: for each block (class), $data$ contains the $x$ and $y$ intervals, where the joints are located, e.g.: $data[0] = [0, 32, 64, 128]$ if $joint_{0}$ is in the interval $[0, 32]$/$[64,128]$ in the $x$/$y$ axis, respectively.
\item In \textit{Line 15 to 17}, $classes$ is a dictionary, where the key is the joint number and the value is a zero vector. To fill in the latter, we iterate over all the $(key, value)$ pairs in $data$ to evaluate the boolean expression in \textit{Line 19}. The latter creates a boolean vector of size $21$ such that each value is \textit{True/False} if the joint is inside/outside the the $x$/$y$ intervals.
\item \textit{Line 20 to 24}: we get the one-hot vector of each joint by assigning the position of the correct class in the zero-vector to 1.
\item From \textit{Line 27 to 30}, $output$ is an array of size $21$, where we get the class of each joint using the \textit{Argmax} function on the one-hot vectors.
\end{itemize}

We follow the same analogy in Algorithm.~\ref{alg:labels3d} to create the ground truth classes of each joint in the 3D space.

\begin{algorithm}
\SetKwInOut{Input}{Input}
\SetKwInOut{Output}{Output}
\underline{function CreateClasses2D} $(2D, splits, size)$\;
\Input{$(x, y)$ points, splits number and image size}
\Output{An array containing the class of each 2D joint}

$block \leftarrow int(size / splits)$\;
$parts, data, output$  $\leftarrow []$\;
$classes \leftarrow$ \{\}\;
$k \leftarrow 0$\;

\For{$i$ \textbf{in} $range(0,size+block,block)$}
{
        $parts[i]$ $\leftarrow i$\;
}
\For{$j$=$0$:$split$}
{
	\For{$i$=$0$:$split$}
	{
		$data[k]$ $\leftarrow$ [$parts[j]$, $parts[j+1]$, $parts[i]$, $parts[i+1]$]\;
		$k \leftarrow k + 1$ \;
	}
}
\For{$i$=$0$:$21$}
{
	$classes[i] \leftarrow [0, ..., 0]$\;
}
\For{$key$, $(a_x,b_x,a_y,b_y)$ \textbf{in} $data.items()$}
{
	$r \leftarrow$ $(a_x <= 2D[:,0])$ \& $(b_x >= 2D[:,0])$ \& $(a_y <= 2D[:,1])$ \& $(b_y >= 2D[:,1])$ \;
	\For{$i$=$0$:$21$}
	{
		\If{$r[i] = True$}
		{
			$classes[i][key] \leftarrow 1$\;
		}
	}
}

$i \leftarrow 0$\;
\For{$key, val$ \textbf{in} $classes.items()$}
{
	$output[i] \leftarrow argmax(val)$\;
	$i \leftarrow i + 1$\;
}
return $output$\;
\caption{Creation of the 2D ground truth classes of each joint in a cropped hand.}
\label{alg:labels2d}
\end{algorithm}

\begin{algorithm}
\SetKwInOut{Input}{Input}
\SetKwInOut{Output}{Output}
\underline{function Create3DClasses} $(3D, splits)$\;
\Input{$(x, y, z)$ points and splits number}
\Output{An array containing the class of each 3D joint}

$start_{x\_3d},start_{y\_3d}, start_{z\_3d}$ $\leftarrow$ $min(3D[:, 0])$,$min(3D[:, 1])$,$min(3D[:, 2])$\;

$end_{x\_3d},end_{y\_3d},end_{z\_3d}$ $\leftarrow$ $max(3D[:, 0])$,$max(3D[:, 1])$,$max(3D[:, 2])$\;

$interval_{x\_3d} \leftarrow end_{x\_3d} - start_{x\_3d}$\;
$interval_{y\_3d} \leftarrow end_{y\_3d} - start_{y\_3d}$\;
$interval_{z\_3d} \leftarrow end_{z\_3d} - start_{z\_3d}$\;

$step_x \leftarrow float(interval_{x\_3d} / splits)$\;
$step_y \leftarrow float(interval_{y\_3d} / splits)$\;
$step_z \leftarrow float(interval_{z\_3d} / splits)$\;

$parts_x, parts_y, parts_z$  $\leftarrow []$\;

$data, output$  $\leftarrow []$\;
$classes$  $\leftarrow \{\}$\;
$k$  $\leftarrow 0$\;

$range_x$ = $arange(start_{x\_3d}, end_{x\_3d} + step_x, step_x)$\;
$range_y$ = $arange(start_{y\_3d}, end_{y\_3d} + step_y, step_y)$\;
$range_z$ = $arange(start_{z\_3d}, end_{z\_3d} + step_z, step_z)$\;

\For{$i,j,k$ \textbf{in} $range_x,range_y,range_z$}
{
	$parts_{x[i]}$ $\leftarrow$ i, $parts_{y[j]}$ $\leftarrow$ j, $parts_{z[k]}$ $\leftarrow$ k\;
}

\For{$i$=$0$:$21$}
{
	$classes[i] \leftarrow [0, ..., 0]$\;
}

\For{$j$=$0$:$n\_splits$}
{
	\For{$i$=$0$:$n\_splits$}
	{
		\For{$p$=$0$:$n\_splits$}
		{
			$data[k]$ $\leftarrow$ $[parts\_x[j], parts\_x[j+1], parts\_y[i], parts\_y[i+1], parts\_z[p], parts\_z[p+1]]$\;
			$k \leftarrow k + 1$\;
		}
	}
}

\For{$key$, $(a_x,b_x,a_y,b_y,a_z,b_z)$ \textbf{in} $data.items()$}
{
	$r \leftarrow$ $(a_x <= 3D[:,0])$ \& $(b_x >= 3D[:,0])$ \& $(a_y <= 3D[:,1])$ \& $(b_y >= 3D[:,1])$ \& $(a_z <= 3D[:,2])$ \& $(b_z >= 3D[:,2])$\;
	\For{$i$=$0$:$21$}
	{
		\If{$r[i] = True$}
		{
			$classes[i][key] \leftarrow 1$\;
		}
	}
}

$i \leftarrow 0$\;
\For{$key, val$ \textbf{in} $classes.items()$}
{
	$output[i] \leftarrow argmax(val)$\;
	$i \leftarrow i + 1$\;
}

return $output$\;
\caption{Creation of the 3D ground truth classes of each joint in a cropped hand.}
\label{alg:labels3d}
\end{algorithm}

\end{document}